\documentclass[10pt,twocolumn,letterpaper]{article}

\usepackage[pagenumbers]{cvpr} 

\usepackage{amsmath}
\usepackage{multirow}
\usepackage{floatrow}
\usepackage{caption}
\newfloatcommand{capbtabbox}{table}[][\FBwidth]

\definecolor{cvprblue}{rgb}{0.21,0.49,0.74}
\usepackage[pagebackref,breaklinks,colorlinks,allcolors=cvprblue]{hyperref}

\def\paperID{xxxx} 
\def\confName{xxxx}
\def\confYear{xxxx}

\title{SEP: Self-Enhanced Prompt Tuning for Visual-Languge Model}

\author{Hantao Yao\\
\and
Rui Zhang\\
\and
Lu Yu\\
\and
Yongdong Zhang\\
\and
Changsheng Xu\\
}

\author{Hantao Yao$^{1}$, Rui Zhang$^{2}$, Lu Yu$^{3}$, Yongdong Zhang$^{4}$, Changsheng Xu$^{1}$\\
{\small $^{1}$ Institute of Automation, Chinese Academy of Sciences; $^{2}$ Institute of Computing Technology, Chinese Academy of Sciences}\\
{\small $^{3}$ Tianjin University of Technology; $^{4}$ University of Science and Technology of China}\\
{\tt\small hantao.yao@nlpr.ia.ac.cn}
}

\begin{document}
\maketitle
\begin{abstract}
Prompt tuning based on Context Optimization (CoOp) effectively adapts visual-language models (VLMs) to downstream tasks by inferring additional learnable prompt tokens. 
However, these tokens are less discriminative as they are independent of the pre-trained tokens and fail to capture input-specific knowledge, such as class-aware textual or instance-aware visual knowledge. 
Leveraging the discriminative and generalization capabilities inherent in pre-trained tokens, we introduce a novel approach named Self-Enhanced Prompt Tuning (SEP). 
The core principle of SEP involves adapting the learnable prompt tokens at each encoder layer from the corresponding self-pretrained tokens, thereby explicitly incorporating discriminative prior knowledge to enhance both textual-level and visual-level embeddings. 
Furthermore, SEP's self-enhanced tokens not only boost discrimination but also mitigate domain shifts in unseen domains, enhancing generalization. 
In practice, SEP selects several representative tokens from all pre-trained tokens for each input data at every layer of the text/visual encoders. 
Subsequently, a Token Fusion Module (TFM) is introduced to generate a self-enhanced token by merging these representative tokens with the learnable tokens using a cross-attention mechanism. 
This self-enhanced token is then concatenated with all pre-trained tokens, serving as input for subsequent encoder layers to produce the relevant embeddings. 
Comprehensive evaluations across various benchmarks and tasks confirm SEP's efficacy in prompt tuning.
Code: \href{Code}{https://github.com/htyao89/SEP/}.
\end{abstract}

\section{Introduction}
Recent studies demonstrate that the Visual Language Model (VLM) possesses a robust generalization capability in a variety of downstream tasks~\cite{abs-2204-14198,RadfordKHRGASAM21}. 
However, the direct application of VLMs in scenarios with limited images is challenging because of their extensive parameters and the substantial image datasets required for training. 
Prompt tuning has recently emerged as an effective and straightforward method for leveraging the general knowledge of VLMs in specialized computer vision and machine learning tasks such as zero-shot learning, domain adaptation, domain generalization, and few-shot learning.

\begin{figure}
  \centering
   \includegraphics[width=1.0\linewidth]{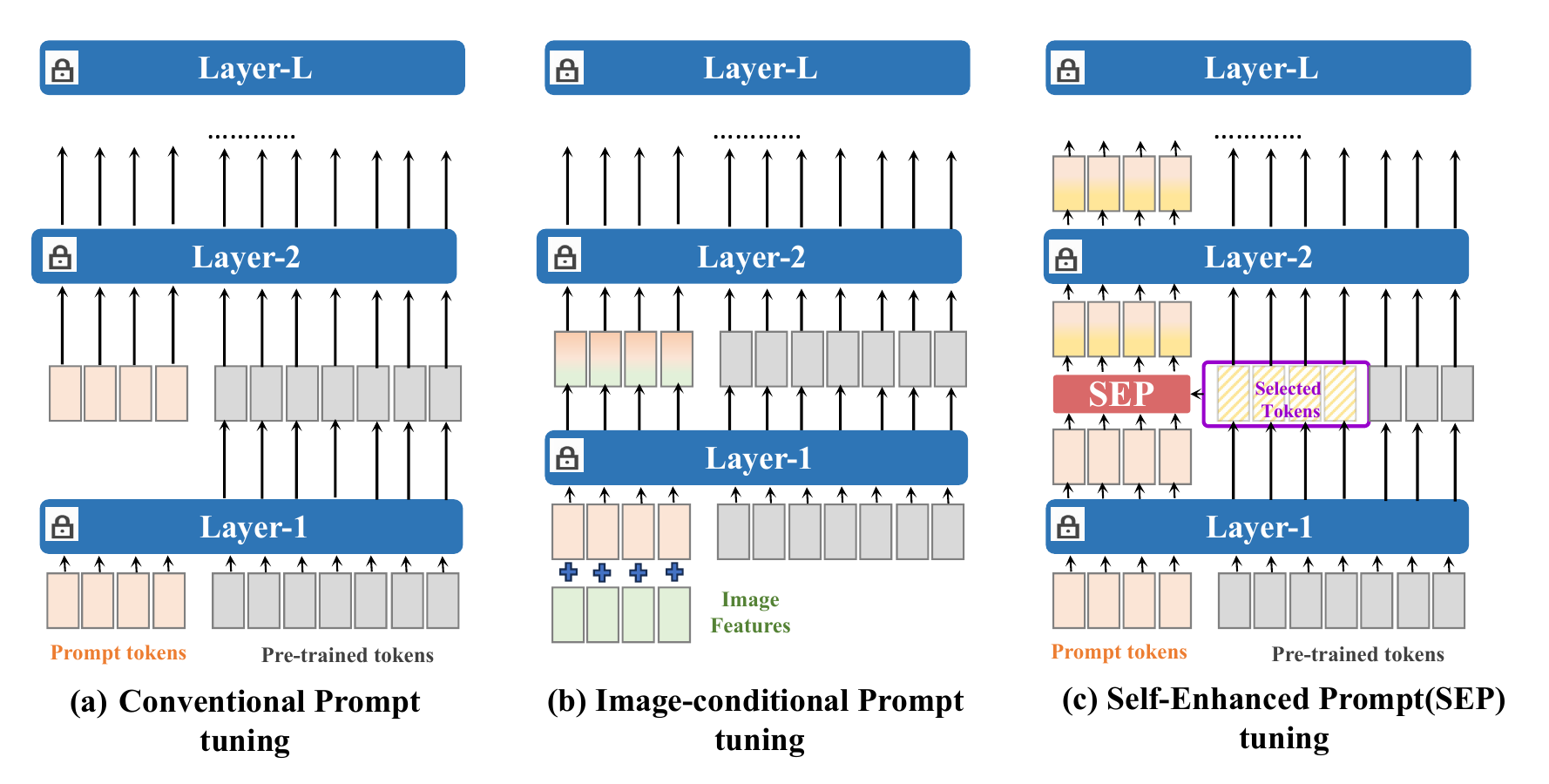}
   \caption{Comparison with existing framework. (a) Conventional input-irrelevant prompt tuning; (b) Image-conditional prompt tuning; (c) Self-enhanced prompt tuning by injecting the discriminative and generalizable knowledge contained in the frozen tokens.}
   \label{fig:motivaiton}
\vspace{-1.0em}
\end{figure} 

Prompt tuning combines additional learnable tokens with pre-trained ones to generate a discriminative embedding, termed Context Optimization (CoOp)~\cite{ZhouYLL22}. 
Recent advances in CoOp-based methods~\cite{ZhouYLL22,YaoZX23,0002YSLR023,BlackBox,KAPT23,DAPT23,RPO23,MAPLE23,DBLP:journals/corr/abs-2303-15234,PromptReg23} aim to develop input-relevant prompt tokens to enhance the discriminative power of embeddings, as illustrated in Figure~\ref{fig:motivaiton}(a). 
The term `input-irrelevant prompt' refers to a prompt initialized random that does not depend on specific input-related knowledge. 
However, since these prompts are derived from the training domain, they may exhibit domain shifts from unseen domains, potentially degrading performance. 
In addition, image-conditional prompts, which incorporate image features into the learnable tokens, have been proposed in~\cite{ZhouYL022,abs-2210-07225} (see Figure~\ref{fig:motivaiton}(b)). 
Despite their specificity, the image-conditional prompts have a limited ability to enhance the class-level classifier. 
In summary, while existing CoOp-based methods generate learnable tokens that are independent of input knowledge, they struggle to utilize essential input-aware prior knowledge, such as instance-aware visual knowledge or class-aware textual knowledge. 
Therefore, how to incorporate discriminative and generalizable input-aware knowledge into learnable prompts is crucial to enhancing their effectiveness for prompt tuning.

Existing methods have demonstrated that the frozen CLIP, trained with a large scale of images, possesses robust generalization capabilities for new classes or images. 
For example, CLIP exhibits remarkable performance in zero-shot learning and domain generalization, indicating that frozen tokens corresponding to each input contain essential generalizable knowledge. 
In addition, these frozen tokens can be used to improve the discriminative and generalization of learnable prompts, resulting in prompts that contain the prior discriminative knowledge related to each input. 
To better mine more useful knowledge, several representative tokens are selected from the pool of all frozen tokens. 
These representative tokens are then integrated with the tokens corresponding to the learnable prompt to generate the input-aware prompt. 
This integration leverages the knowledge embedded in the frozen tokens, endowing the resultant input-aware tokens with several advantages. 
Unlike the input-irrelevant prompt, input-aware prompts are related to each input due to their incorporation of specific knowledge from the frozen token, thereby enhancing their generalizability.
Furthermore, discriminative knowledge within the representative tokens injects the input-aware prompt with discriminative capabilities.
In summary, by associating pre-trained tokens with learnable prompts, a robust input-aware token is generated, significantly boosting both the generalization and discriminative capacities of prompt tuning.

In this work, we propose a novel approach named Self-Enhanced Prompt Tuning (SEP), as depicted in Figure~\ref{fig:sep}. 
SEP introduces a Token-Fusion Module (TFM) to incorporate the essential knowledge from pre-trained tokens into input-aware tokens. 
Specifically, TFM selects a subset of highly discriminative pre-trained tokens for each input from the middle layers. 
It then employs a multi-head attention mechanism to integrate these selected tokens with the corresponding prompt-related tokens for generating a self-enhanced prompt token. 
Subsequently, the enhanced token is concatenated with the pre-trained tokens and processed by subsequent encoder layers. 
In particular, TFM can be seamlessly integrated into the middle layers of both visual and textual encoders in the CLIP.

The proposed Self-Enhanced Prompt Tuning (SEP) significantly improves discriminativeness and generalizability for downstream tasks by injecting the essential knowledge contained in the pre-trained tokens. 
The effectiveness of SEP is validated in several tasks and benchmarks: base-to-new settings, domain generalization, cross-dataset generalization, and few-shot learning. 
The major contributions of SEP include: 1) the introduction of an innovative Self-Enhanced Prompt tuning mechanism, which integrates self-enhanced prompt tokens generated by the Token-Fusion Module (TFM) into visual/textual encoders; 
2) we claim that the injection of prior discriminative knowledge from pre-trained tokens into learnable prompt tokens can enhance the  discriminative and generalizable of the generated embeddings.

\section{Related Works}
\label{sec:related}
\textbf{Vision-Language Models:}
Recently, Vision-Language Models (VLMs)~\cite{RadfordKHRGASAM21,abs-2204-14198}, which utilize both visual and textual encoders, have been trained on a large-scale dataset of image-text pairs, demonstrating significant generalization and discrimination capabilities. 
To further enhance the descriptive power of VLMs, improvements have been made in several areas: (1) employing more robust text or visual encoders~\cite{VaswaniSPUJGKP17,zhai2022scaling,li2023blip}; (2) integrating visual and textual knowledge more deeply~\cite{li2022blip,singh2022flava}; and (3) increasing the volume of images used~\cite{RadfordKHRGASAM21,JiaYXCPPLSLD21,schuhmann2021laion,schuhmann2022laion}. To enhance the diversity of text descriptions, Masked Language Modeling (MLM)~\cite{KimSK21,LuBPL19} has been employed, which randomly omits words in the text to aid in representation learning. 
In contrast to MLM, masked autoencoder-based methods~\cite{HeCXLDG22} have been introduced to improve descriptive capabilities by randomly masking image patches. 
Among the VLM frameworks, CLIP stands out as a straightforward yet effective model that uses contrastive loss to train separate visual and text encoders based on 400 million image-text pairs. 
Given its robust generalization, many CoOp-based methods have been developed using CLIP to adapt pre-trained VLMs for downstream tasks. 
In this work, we have implemented a prompt-tuning strategy to generate task-specific embeddings.

\textbf{Prompt Tuning:}
To adapt pre-trained Vision-Language Models (VLMs) to downstream tasks, various strategies involving prompt tuning have been employed~\cite{abs-2210-09263,BlackBox,RPO23,abs-2210-07225,RadfordKHRGASAM21,DBLP:journals/corr/abs-2303-06571}. 
Initially, CLIP\cite{RadfordKHRGASAM21} uses a hand-crafted template `a photo of a [CLASS]' for zero-shot prediction. 
However, these hand-crafted prompts often fail to adequately capture the nuances of downstream tasks, leading to the development of textual prompt tuning strategies that use learnable textual tokens alongside class tokens to enhance textual embeddings.
For example, the hand-crafted prompts are replaced with the learnable prompt tokens in Context Optimization(CoOp)~\cite{ZhouYLL22}.
Moreover, Conditional Context Optimization(CoCoOp)~\cite{ZhouYL022} and VPT~\cite{abs-2210-07225} generate an image-conditional prompt fusing the image feature and the learnable textual prompt to boost generalization in the unseen domain.
To better use the essential knowledge contained in frozen CLIP, Knowlege-Guided Context Optimization(KgCoOp)~\cite{YaoZX23}, ProGrad~\cite{abs-2205-14865}, and Prompt Regularization(ProReg)~\cite{PromptReg23} all construct a novel constraint term by constraining the consistency between the learnable knowledge and the essential general knowledge.
Unlike the above methods, which consider textual prompts, Ensembling Context Optimization(ECO)~\cite{ECO23} employs prompt ensembling to combine multiple prompts.
Furthermore, ProDA~\cite{proda} considers the prompt's prior distribution learning, and Distribution-Aware Prompt Tuning (DAPT)~\cite{DAPT23} optimizes the learnable prompt by maximizing inter-dispersion.
Knowlege-Aware Prompt Tuning(KAPT)~\cite{KAPT23} employs external knowledge to boost generalization on unseen classes.
PLOT~\cite{0002YSLR023} applies optimal transport to match the vision and text modalities for generating the discriminative and visual-aligned local textual prompt tokens.
Besides the textual prompt tuning, Multi-modal Prompt Learning (MaPLe)~\cite{MAPLE23} and PromptSRC~\cite{DBLP:journals/corr/abs-2307-06948} conduct the visual-textual prompt tuning by jointly conducting the prompt tuning on the visual and text encoders.
Multitask Vision-Language Prompt Tuning (MVLPT)~\cite{DBLP:journals/corr/abs-2211-11720} integrates cross-task knowledge, enhancing the versatility of prompt tuning in VLMs.
Finally, DenseCLIP~\cite{RaoZ0TZH0L22} and CLIP-Adapter~\cite{abs-2110-04544}  introduce context-aware strategies and adapters to refine the embeddings for specific tasks, further illustrating the dynamic evolution of prompt tuning in the field of vision-language models.
Recently, GraphAdaper~\cite{DBLP:conf/nips/0082LLB0W23} is proposed to fuse visual and textual knowledge with two Graph-based Adapters, and the Class-aware prompt is proposed in TCP~\cite{yao2023tcp} to inject class-level knowledge into the prompt.

Existing methods generate two types of input-irrelevant and image-conditional tokens, exhibiting limited capability to capture specific knowledge. 
To address this issue, we introduce a novel Self-Enhanced Prompt Tuning, which utilizes the Token Fusion Module (TFM) to integrate discriminative knowledge from self-pretrained tokens into the enhanced prompts, thereby enriching the prompt's contextual relevance.
Unlike traditional CoOp-based methods, our technique distinctively generates prompt tokens by leveraging these self-pretrained tokens. 
Our evaluations across various tasks and benchmarks demonstrate that this integration significantly boosts the discriminative power of the prompt tuning process, confirming the efficacy of the proposed SEP.

\section{Methodolgy}

\subsection{ Preliminaries}
The frozen Contrastive Language-Image Pre-training(CLIP) trained with a large number of images exhibits robust generalization capacity to new classes or images, \emph{e.g.,} CLIP shows impressive performance on zero-shot learning and domain generalization tasks,
Therefore, existing prompt tuning based on Context Optimization(CoOp) is proposed based on CLIP.
Given an image along with all class names, CLIP extracts the visual and text embeddings with the visual and text encoders, and the contrast loss is calculated to align those two embeddings.
To effectively adapt CLIP for the downstream task, CLIP applies the hand-crafted template "a photo of a \{\}” to extract the general class-level textual embedding, defined as $\mathbf{W}^{clip}=\{\mathbf{w}^{clip}_i\}_{i=1}^{{N}_c}$, where $\mathbf{w}^{clip}_i$ is the textual embedding of $i$-th class, and $N_c$ is the number of classes.
Given the `class-name' of $i$-th class, Word Embedded $e(\cdot)$ first embeds the hand-crafted description into a vectorized textual tokens: $\mathbf{t}^{clip}_i=e$("a photo of a \{class-name\}"). 
After that, Text Encoder $\psi$ maps the vectorized textual tokens $\mathbf{t}^{clip}_i$ to the class-level embedding: $\mathbf{w}^{clip}_i=\psi(\mathbf{t}^{clip}_i)$.

In addition, Context Optimization (CoOp) replaces hand-crafted text tokens with learnable text tokens $\mathbb{T}=\{\mathbf{t}_1, \mathbf{t}_2,...,\mathbf{t}_{M}\}$, where $M$ is the length of the tokens.
Similarly to CLIP, the corresponding class token $\mathbf{c}_i$ is concatenated with the learnable tokens $\mathbb{T}$ to generate the textual token $\mathbf{t}^{coop}_i=\{\mathbf{t}_1, \mathbf{t}_2,...,\mathbf{t}_{M},\mathbf{c}_i\}$.
Next, textual embedding $\mathbf{w}^{coop}_{i}$ is obtained by feeding textual tokens $\mathbf{t}^{coop}_i$ into the Text Encoder $\psi$, \emph{i.e.}, $\mathbf{w}^{coop}_i=\psi(\mathbf{t}^{coop}_i)$.
Finally, textual embeddings of all classes are defined as $\mathbf{W}^{coop}=\{\mathbf{w}^{coop}_i\}_{i=1}^{{N}_c}$.
CoOp infers the learnable textual tokens $\mathbb{T}$ by minimizing the contrastive loss between the image's embedding $\mathbf{f}$ and its class embedding $\mathbf{w}^{coop}_{y}$:
\begin{equation}
\mathcal{L}_{ce}=\frac{1}{N}\sum_{(\mathbf{f},y)\in \mathcal{D}_{s}}\frac{\exp(d(\mathbf{f},\mathbf{w}^{coop}_y)/\tau)}{\sum_{i=1}^{N_c}\exp(d(\mathbf{f},\mathbf{w}^{coop}_i)/\tau)},
\label{eq:coop}
\end{equation}
where $\mathcal{D}_{s}$ is the seen dataset, and $d(\cdot)$ is the cosine distance.
$\tau$ is a temperature factor defined in CLIP, and $N$ is the number of training images.

As the generated textual embedding has a good generalization ability for the novel classes, KgCoOp further adds an efficient consistency $\mathcal{L}_{kg}$ between the generated embedding $\mathbf{W}^{coop}$ and the general embedding $\mathbf{W}^{clip}$,
\begin{equation}
\mathcal{L}_{kg}=||\mathbf{W}^{clip}-\mathbf{W}^{coop}||_{2}^{2}.
\end{equation}

Therefore, a robust objective for prompt tuning is:
\begin{equation}
\mathcal{L} = \mathcal{L}_{ce}+\omega\mathcal{L}_{kg},
\label{eq:L}
\end{equation}
where $\omega$ is set as 8.0 same as in KgCoOp~\cite{YaoZX23}.

\begin{figure*}
  \centering
   \includegraphics[width=0.75\linewidth]{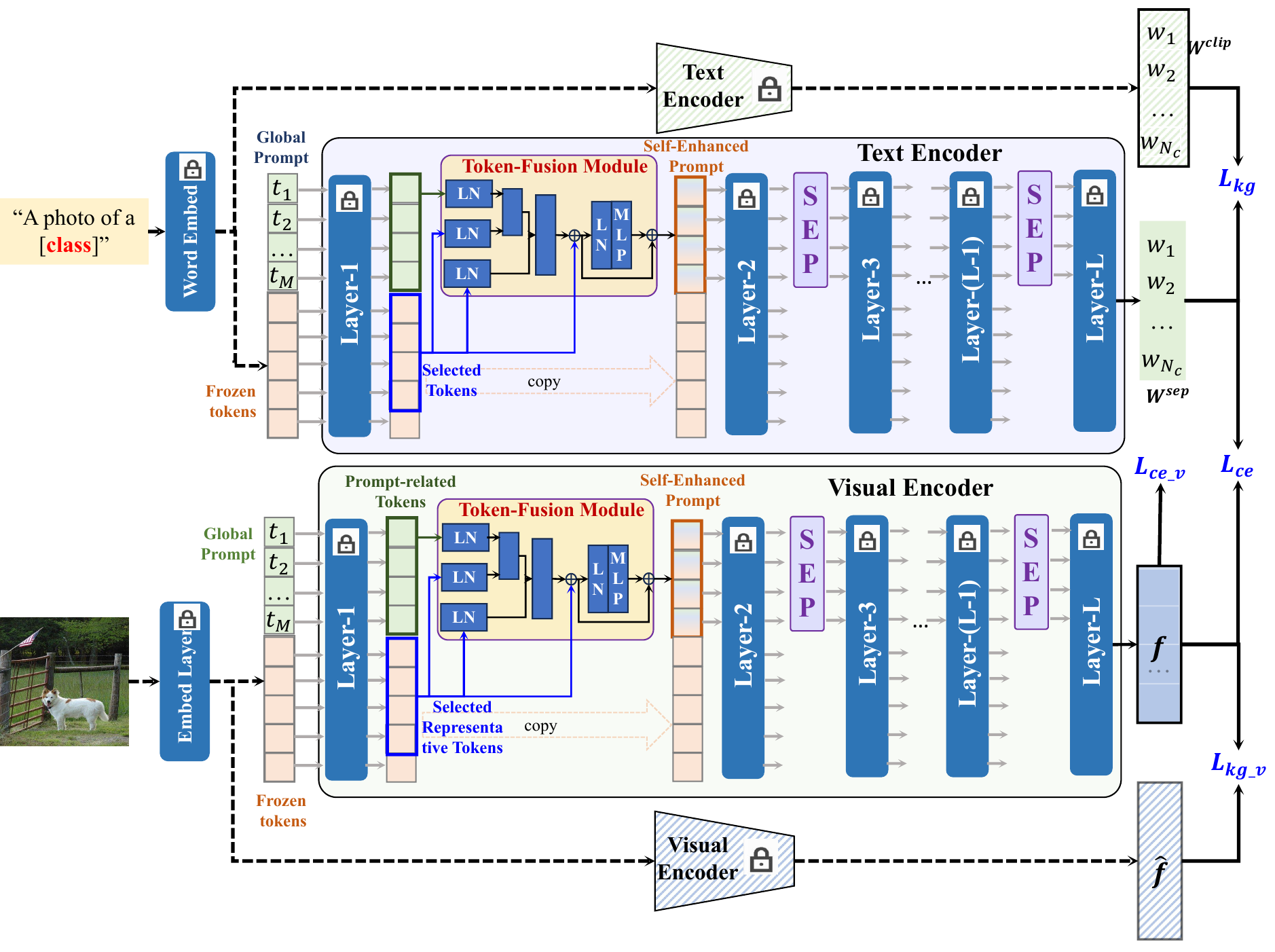}
   \caption{The framework of the proposed Self-Enhanced Prompt tuning. The Token-Fusion Module is used to integrate the pre-trained tokens and the prompt-related tokens for generating the self-enhanced prompt.}
   \label{fig:sep}
\vspace{-1.0em}
\end{figure*} 

\subsection{Self-Enhanced Prompt Tuning}
Most existing CoOp-based methods infer input-irrelevant and image-conditional prompt tokens for prompt tuning. 
However, the input-irrelevant prompts are typically derived from the seen domain, inherently biasing them from the unseen domain and consequently degrading performance. 
Additionally, image-conditional prompt tokens that incorporate image-specific knowledge are less effective in mitigating domain shifts at the class level.
A significant limitation of current approaches is their inability to encapsulate prior essential knowledge; for instance, neither visual nor textual prompts adequately capture instance-aware visual knowledge or class-aware textual knowledge within the visual or text encoders, respectively. 
Recognizing that essential knowledge generated by the frozen CLIP model exhibits high discriminative and generative capabilities, a reasonable motivation is to leverage this inherent knowledge to boost the prompt tuning. 
By integrating essential knowledge from frozen tokens into learnable prompts, an enhanced token that is aware of the input can be generated that offers distinct advantages.
Firstly, input-aware prompt tokens inherit discriminative knowledge from frozen tokens, enhancing the discriminative capacity of generated embeddings across both seen and unseen domains. 
Secondly, these tokens possess enhanced generability for unseen domains because the frozen tokens from these domains provide crucial insights that guide the embedding generation during testing.
In summary, by fusing frozen tokens with learnable prompts, we can formulate robust input-aware tokens that significantly boost the generalization and discriminative capabilities of prompt tuning. 
Therefore, we introduce a novel Self-Enhanced Prompt Tuning (SEP), as depicted in Figure~\ref{fig:sep}. 
Notably, SEP can be easily applied to both Visual and Text Encoders, and we introduce the application of SEP on the Visual Encoder as an illustrative example.

Given the batch images $\mathbf{X}$ consisting of $N_b$ images, we first apply the pre-trained Embedding Layer $\mathcal{E}(\cdot)$ to extract the frozen image tokens $\mathbf{E}\in \mathbb{R}^{L_e\times N_b \times D_v}$, where $L_e=193$ and $D_v=768$ are tokens' length and dimension for CLIP with ViT-B/16.
Similarly to multi-modal prompt tuning~\cite{MAPLE23}, a global visual prompt $\mathbf{P}_g\in \mathbb{R}^{L_v\times D_v}$ is first combined with the frozen image tokens $\mathbf{E}$, where $L_v$ is the length of the global visual prompt.
Basically, the extended image tokens are $\mathbf{V}_{0}=[\mathbf{E},\mathbf{P}_g]\in\mathbb{R}^{(L_e+L_v)\times N_b \times D_v}$.
Subsequently, the extended image tokens $\mathbf{V}_{0}$ are fed into the Visual Encoder $\theta$ for embedding.
Specifically, the first layer of the Visual Encoder $\theta_{1}$ projects the image tokens $\mathbf{V}_{0}$ into the middle token $\mathbf{V}_{1}$,
\begin{equation}
\mathbf{V}_{1}=\theta_{1}(\mathbf{V}_{0}), \mathbf{V}_{0}=[\mathbf{E},\mathbf{P}_g].
\end{equation}

\textbf{Self-Enhanced Prompt tuning:} Once the token $\mathbf{V}_{1}$ is obtained, Self-Enhanced Prompt(SEP) tuning is proposed to generate the corresponding self-enhanced tokens $\mathbf{\hat{V}_1}$ with Token-Fusion Module(TFM).
Note that the image token $\mathbf{V}_{1}=[\mathbf{V}^{v}_{1},\mathbf{V}^{p}_{1}]$ can be split into two components: the pre-trained visual tokens $\mathbf{V}^{v}_{1}=\mathbf{V}_{1}[1:L_e,:,:]$, and the prompt-related tokens $\mathbf{V}^{p}_{1}=\mathbf{V}_{1}[L_e+1:L_e+L_v,:,:]$\footnote{For the visual prompt tuning, the prompt-related tokens are always inserted after the original visual tokens.}.
The pretrained visual tokens $\mathbf{V}^{v}_{1}$ are mostly related to the pre-trained image tokens $\mathbf{E}$ that contain the essential discriminative and generalization knowledge, while the prompt-related tokens are most related to the additional learnable prompt to capture the transfer knowledge.

Unlike existing methods that simply fed the prompt-related tokens $\mathbf{V}^{p}_{1}$ into the next layer, Token-Fusion Module(TFM) ($\mathcal{T}^{v}$) is proposed to generate the self-enhanced token $\mathbf{\hat{V}}^{p}_{1}$ by associating the pre-trained visual tokens $\mathbf{V}^{v}_{1}$ and the prompt-related tokens $\mathbf{V}^{p}_{1}$,
\begin{equation}
\mathbf{\hat{V}}^{p}_{1}=\mathcal{T}^{v}(\mathbf{V}^{v}_{1},\mathbf{V}^{p}_{1}).
\end{equation}

The generated self-enhanced token $\mathbf{\hat{V}}^{p}_{1}$ has inherited the essential knowledge from the pre-trained visual tokens $\mathbf{V}^{v}_{1}$, which means that the generated embedding has enough discriminative ability.
After that, the self-enhanced token $\mathbf{\hat{V}}^{p}_{1}$ is combined with the pre-trained visual tokens $\mathbf{V}^{v}_{1}$ to generate the new visual tokens $\mathbf{\hat{V}}_{1}=[\mathbf{V}^{v}_{1},\mathbf{\hat{V}}^{p}_{1}]$, fed into the next layers for embedding.

\begin{table*}[htb]
\caption{Comparion on the base-to-new generalization. `*' denotes the results are based our re-implemented.}
\vspace{-1.0em}
\label{tab:base2new}
\resizebox{\linewidth}{!}{
\begin{tabular}{lc|ccccccccccccc|c}
\hline
                    & Sets & CoOp* & CoCoOp & DAPT* & ProGrad*       & ProDA          & KgCoOp & RPO            & PLOT* & LFA   & MaPLe          & DePT  & PromptSRC      & TCP            & SEP            \\ \hline
\multirow{3}{*}{\rotatebox{90}{\small Average}}    & Base & 82.38 & 80.47  & 83.18 & 82.48          & 81.56          & 80.73  & 81.13          & 83.98 & 83.62 & 82.28          & 83.62 & 84.26          & 84.13          & \textbf{85.98} \\
                            & New  & 67.96 & 71.69  & 69.27 & 70.75          & 72.30          & 73.6   & 75.00          & 71.72 & 74.56 & 75.14          & 75.04 & 76.10          & 75.36          & \textbf{76.49} \\
                            & H    & 74.48 & 75.83  & 75.59 & 76.16          & 76.65          & 77.0   & 77.78          & 77.37 & 78.83 & 78.55          & 79.10 & 79.97          & 79.51          & \textbf{80.96} \\ \hline
\multirow{3}{*}{\rotatebox{90}{\small ImageNet}}   & Base & 76.46 & 75.98  & 76.83 & 77.02          & 75.40          & 75.83  & 76.60          & 77.30 & 76.89 & 76.66          & 77.03 & 77.60          & 77.27          & \textbf{78.10} \\
                            & New  & 66.31 & 70.43  & 69.27 & 66.66          & 70.23          & 69.96  & \textbf{71.57} & 69.87 & 69.36 & 70.54          & 70.13 & 70.73          & 69.87          & 69.55          \\
                            & H    & 71.02 & 73.10  & 72.85 & 71.46          & 72.72          & 72.78  & 74.00          & 73.40 & 72.93 & 73.47          & 73.42 & \textbf{74.01} & 73.38          & 73.58          \\ \hline
\multirow{3}{*}{\rotatebox{90}{\small Caltech}} & Base & 97.80 & 97.96  & 97.83 & 98.02          & 98.27          & 97.72  & 97.97          & 98.53 & 98.41 & 97.74          & 98.30 & 98.10          & 98.23          & \textbf{98.80} \\
                            & New  & 93.27 & 93.81  & 93.07 & 93.89          & 93.23          & 94.39  & 94.37          & 92.80 & 93.93 & 94.36          & 94.60 & 94.03          & \textbf{94.67} & 94.53          \\
                            & H    & 95.48 & 95.84  & 95.39 & 95.91          & 95.68          & 96.03  & 96.03          & 95.58 & 96.13 & 96.02          & 96.41 & 96.02          & 96.42          & \textbf{96.62} \\ \hline
\multirow{3}{*}{\rotatebox{90}{\small Pets}} & Base & 94.47 & 95.20  & 95.00 & 95.07          & \textbf{95.43} & 94.65  & 94.63          & 94.50 & 95.13 & 95.43          & 94.33 & 95.33          & 94.67          & 95.37          \\
                            & New  & 96.00 & 97.69  & 95.83 & \textbf{97.63} & 97.83          & 97.76  & 97.50          & 96.83 & 96.23 & 97.76          & 97.23 & 97.30          & 97.20          & 97.40          \\
                            & H    & 95.23 & 96.43  & 95.41 & 96.33          & \textbf{96.62} & 96.18  & 96.05          & 95.65 & 95.68 & 96.58          & 95.76 & 96.30          & 95.92          & 96.37          \\ \hline
\multirow{3}{*}{\rotatebox{90}{\small Cars}}       & Base & 75.67 & 70.49  & 75.80 & 77.68          & 74.70          & 71.76  & 73.87          & 79.07 & 76.32 & 72.94          & 79.13 & 78.27          & 80.80          & \textbf{83.77} \\
                            & New  & 67.53 & 73.59  & 63.93 & 68.63          & 71.20          & 75.04  & \textbf{75.53} & 74.80 & 74.88 & 74.00          & 75.47 & 74.97          & 74.13          & 75.43          \\
                            & H    & 71.37 & 72.01  & 69.36 & 72.88          & 72.91          & 73.36  & 74.69          & 76.88 & 75.59 & 73.47          & 77.26 & 76.58          & 77.32          & \textbf{79.38} \\ \hline
\multirow{3}{*}{\rotatebox{90}{\small Flowers}}    & Base & 97.27 & 94.87  & 96.97 & 95.54          & 97.70          & 95.00  & 94.13          & 97.93 & 97.34 & 95.92          & 98.00 & 98.07          & 97.73          & \textbf{98.17} \\
                            & New  & 67.13 & 71.75  & 60.90 & 71.87          & 68.68          & 74.73  & 76.60          & 73.53 & 75.44 & 72.46          & 76.37 & 76.50          & 75.57          & \textbf{76.63} \\
                            & H    & 79.44 & 81.71  & 74.81 & 82.03          & 80.66          & 83.65  & 84.50          & 83.99 & 85.00 & 82.56          & 85.84 & 85.95          & 85.23          & \textbf{86.07} \\ \hline
\multirow{3}{*}{\rotatebox{90}{\small Food}}    & Base & 89.37 & 90.70  & 90.37 & 90.37          & 90.30          & 90.5   & 90.33          & 89.80 & 90.52 & \textbf{90.71} & 90.50 & 90.67          & 90.57          & 90.60          \\
                            & New  & 88.77 & 91.29  & 91.30 & 89.59          & 88.57          & 91.7   & 90.83          & 91.37 & 91.48 & \textbf{92.05} & 91.60 & 91.53          & 91.37          & 91.60          \\
                            & H    & 89.07 & 90.99  & 90.83 & 89.98          & 89.43          & 91.09  & 90.58          & 90.58 & 91.00 & \textbf{91.38} & 91.05 & 91.10          & 90.97          & 91.10          \\ \hline
\multirow{3}{*}{\rotatebox{90}{\small Aircraft}}   & Base & 39.67 & 33.41  & 39.97 & 40.54          & 36.90          & 36.21  & 37.33          & 42.13 & 41.48 & 37.44          & 43.20 & 42.73          & 41.97          & \textbf{49.17} \\
                            & New  & 31.23 & 23.71  & 29.80 & 27.57          & 34.13          & 33.55  & 34.20          & 33.73 & 32.29 & 35.61          & 34.83 & \textbf{37.87} & 34.43          & 35.17          \\
                            & H    & 34.95 & 27.74  & 34.14 & 32.82          & 35.46          & 34.83  & 35.70          & 37.46 & 36.31 & 36.50          & 38.57 & 40.15          & 37.83          & \textbf{41.01} \\ \hline
\multirow{3}{*}{\rotatebox{90}{\small SUN397}}     & Base & 80.85 & 79.74  & 80.97 & 81.26          & 78.67          & 80.29  & 80.60          & 82.20 & 82.13 & 80.82          & 82.33 & 82.67          & 82.63          & \textbf{82.77} \\
                            & New  & 68.34 & 76.86  & 76.97 & 74.17          & 76.93          & 76.53  & 77.80          & 73.63 & 77.20 & 78.70          & 77.80 & 78.47          & 78.20          & \textbf{78.80} \\
                            & H    & 74.07 & 78.27  & 78.92 & 77.55          & 77.79          & 78.36  & 79.18          & 77.68 & 79.59 & 79.75          & 80.00 & 80.52          & 80.35          & \textbf{80.74} \\ \hline
\multirow{3}{*}{\rotatebox{90}{\small DTD}}        & Base & 79.97 & 77.01  & 82.23 & 77.35          & 80.67          & 77.55  & 76.70          & 81.97 & 81.29 & 80.36          & 82.20 & 82.37          & 82.77          & \textbf{85.5}  \\
                            & New  & 48.60 & 56.00  & 54.23 & 52.35          & 56.48          & 54.99  & 62.13          & 43.80 & 60.63 & 59.18          & 59.13 & \textbf{62.97} & 58.07          & 61.97          \\
                            & H    & 60.46 & 64.85  & 65.36 & 62.45          & 66.44          & 64.35  & 68.61          & 57.09 & 69.46 & 68.16          & 68.78 & 71.75          & 68.25          & \textbf{71.86} \\ \hline
\multirow{3}{*}{\rotatebox{90}{\small EuroSAT}}    & Base & 90.10 & 87.49  & 94.73 & 90.11          & 83.90          & 85.64  & 86.63          & 93.70 & 93.40 & 94.07          & 89.03 & 92.90          & 91.63          & \textbf{95.3}  \\
                            & New  & 53.00 & 60.04  & 50.33 & 60.89          & 66.00          & 64.34  & 68.97          & 62.67 & 71.24 & 73.23          & 71.07 & 73.90          & 74.73          & \textbf{79.6}  \\
                            & H    & 66.74 & 71.21  & 65.74 & 72.67          & 73.88          & 73.48  & 76.79          & 75.11 & 80.83 & 82.3           & 79.04 & 82.32          & 82.32          & \textbf{86.75} \\ \hline
\multirow{3}{*}{\rotatebox{90}{\small UCF101}}     & Base & 84.53 & 82.33  & 84.30 & 84.33          & 85.23          & 82.89  & 83.67          & 86.60 & 86.97 & 83.00          & 85.80 & 87.10          & 87.13          & \textbf{88.23} \\
                            & New  & 67.37 & 73.45  & 76.33 & 74.94          & 71.97          & 76.67  & 75.43          & 75.90 & 77.48 & 78.66          & 77.23 & 78.80          & \textbf{80.77} & 80.73          \\
                            & H    & 74.98 & 77.67  & 80.12 & 79.35          & 78.04          & 79.65  & 79.34          & 80.90 & 81.95 & 80.77          & 81.29 & 82.74          & 83.83          & \textbf{84.31} \\ \hline
\end{tabular}}
\vspace{-1.0em}
\end{table*}

Overall, given the current token $\mathbf{V}_{l}$ of $l$-th layer($l\ge 1$), SEP with the Token-Fusion Module ($\mathcal{T}^{v}$)can be formulated as follows,
\begin{equation}
\label{eq:TFE}
\mathbf{V}_{l+1}=\theta_{l+1}([\mathbf{V}^{v}_{l},\mathcal{T}_{l}^{v}(\mathbf{V}^{v}_{l},\mathbf{V}^{p}_{l})]),(l\ge 1),
\end{equation}
where $\theta_{l+1}$ is the (l+1)-th layer in Visual Encoder.

\textbf{Token-Fusion Module:} In Eq.~\eqref{eq:TFE}, Token-Fusion Module ($\mathcal{T}^{v}$) is the critical aspect for the proposed SEP.
We next give a detailed description of the Token-Fusion Module(TFM).
Given the pre-trained visual tokens $\mathbf{V}^{v}_{l}\in\mathbb{R}^{L_e\times N_b\times D_v}$,  TFM first selects $L_v$ representive tokens $\mathbf{\hat{V}}_{l}^{v}$from all $L_e$ visual tokens, leading to the same dimension as the prompt-related tokens $\mathbf{V}^{p}_{l}$.
Moreover, the activation score of a feature can represent its discriminative ability, \emph{i.e.,} the element with a higher activation score represents the more discriminative.  
Therefore, we choose the top-$L_v$ tokens with the highest activation score.
Note that the activation score of each token is the average of the feature's value after the square.
Consequently, we can obtain top-$L_v$ tokens $\mathbf{\hat{V}}^{v}_{l}$.
After that, cross-attention is applied to fuse the selected visual tokens $\mathbf{\hat{V}}^{v}_{l}$ and the prompt-related tokens $\mathbf{V}^{p}_{1}$,
\begin{align}
\mathbf{\hat{V}}^{p}_{l}&=\mathcal{T}_{l}^{v}(\mathbf{\hat{V}}^{v}_{l},\mathbf{V}^{p}_{1})\\
				  &=softmax\left(\frac{\mathbf{\hat{V}}^{v}_{l}(\mathbf{{V}}^{p}_{l})^{\top}}{\sqrt{d_k}}\right)\mathbf{\hat{V}}^{v}_{l},
\end{align}
where $\mathbf{\hat{V}}^{p}_{l}$ is the generated self-enhanced token, and $d_k$ is the dimension of keys.
$\mathbf{\hat{V}}^{p}_{l}$ is concated with the frozen visual tokens $\mathbf{V}^{v}_{l}$ to generate the new visual tokens $\mathbf{\hat{V}}_{1}=[\mathbf{V}^{v}_{1},\mathbf{\hat{V}}^{p}_{1}]$, fed into the next layers for embedding.

Importantly, the proposed SEP can also be easily applied to the Text Encoder for textual prompt tuning. 
When performing SEP on the Text Encoder, the representative tokens are selected by the front top-$L_t$ tokens containing the discriminative knowledge based on the hand-crafted templates "a photo of a \{class-name\}".

\textbf{Objective:}
Given the image $x$, labels, and the class names, we first use the frozen CLIP to extract the frozen text-level classifier $\mathbf{W}^{clip}$ and the visual embedding $\mathbf{f}$.
By performing the proposed SEP on the Visual Encoder and the Textual Encoder, we can generate the enhanced text-level classifier $\mathbf{W}^{sep}$ and the enhanced visual embedding $\mathbf{\hat{f}}$.
To increase the discrimination of the enhanced visual embedding, a cross-entropy loss $\mathcal{L}_{ce\_v}$ is conducted on the visual embedding $\mathbf{\hat{f}}$.
Inspired by KgCoOp~\cite{YaoZX23}, we also conduct the consistency constraint $\mathcal{L}_{kg\_v}$ between the original visual embedding $\mathbf{f}$ and the enhanced visual embedding $\mathbf{\hat{f}}$:
\begin{equation}
\mathcal{L}_{kg\_v}=||\mathbf{\hat{f}}-\mathbf{f}||^{2}_{2}.
\end{equation}

By considering Eq.~\eqref{eq:L}, the final objective is:
\begin{align}
\mathcal{L}=\mathcal{L}_{ce}(\mathbf{\hat{f}},\mathbf{W}^{sep})+\omega_{t}\mathcal{L}_{kg}(\mathbf{W}^{clip},\mathbf{W}^{sep})\\
+\omega_{v}\mathcal{L}_{kg\_v}(\mathbf{\hat{f}},\mathbf{f})+\mathcal{L}_{ce\_v}(\mathbf{\hat{f}}),
\label{eq:final}
\end{align}
where $\omega_t$ and $\omega_v$ are the weight to balance the effect of the consistency of the textual-level embedding and visual embeddings. 
Similarly to KgCoOp~\cite{YaoZX23}, $\omega_t$ is set as 8.0. The effect of $\omega_v$ is given in the experiment.

\section{Experiments}
\label{sec:exp}

\noindent\textbf{Dataset.}
We conduct the evaluation on the following benchmarks, \emph{i.e.,} ImageNet~\cite{DengDSLL009}, Caltech~\cite{Fei-FeiFP07}, OxfordPets~\cite{ParkhiVZJ12}, StanfordCars~\cite{Krause0DF13}, Flowers~\cite{NilsbackZ08}, Food101~\cite{BossardGG14}, FGVCAircraft~\cite{MajiRKBV13}, EuroSAT~\cite{HelberBDB19}, UCF101~\cite{abs-1212-0402}, DTD~\cite{CimpoiMKMV14}, and SUN397~\cite{XiaoHEOT10}.
Moreover, we use the ImageNet and its variants for domain generalization, \emph{i.e.,} the ImageNet is treated as the source domain; ImageNetV2~\cite{RechtRSS19}, ImageNet-Sketch~\cite{WangGLX19}, ImageNet-A~\cite{GaoZYLGW22} and ImageNet-R~\cite{HendrycksBMKWDD21} are treated as the target domains for evaluation.

\noindent\textbf{Training Details.}
Our implementation is adapted from the public code~\cite{ZhouYLL22,YaoZX23} based on the CLIP with the backbone of ViT-B/16~\cite{DosovitskiyB0WZ21}.
The prompt's length $L_v$ and $L_t$  is set as 4 and 6 for the visual prompt tuning and textual prompt tuning.
The hand-crafted templates initialized with "X X X X X X \{ \}" are used for the textual prompt tuning.
The final performance is averaged over three standard random seeds(1/2/3).
Adam optimizer is applied for optimization with the learning rate of 2.5e-3 and the batch size of 32 up to the training epochs is 50.
All experiments are conducted on RTX 3090.

\noindent\textbf{Baselines.} Recently CoOp-based methods are used for comparison, \emph{e.g.,} CoOp~\cite{ZhouYLL22}, CoCoOp~\cite{ZhouYL022}, ProGrad~\cite{abs-2205-14865}, ProDA~\cite{proda}, KgCoOp~\cite{YaoZX23}, PromptSRC~\cite{DBLP:journals/corr/abs-2307-06948}, MaPLe~\cite{MAPLE23}, LFA~\cite{BlackBox}, DePT~\cite{DBLP:journals/corr/abs-2309-07439}, DAPT~\cite{DAPT23}, PLOT~\cite{0002YSLR023}, TaskRes~\cite{DBLP:conf/cvpr/0012LJ0W23}, RPO~\cite{RPO23},  VPT~\cite{abs-2210-07225}, GraphAdaper~\cite{DBLP:conf/nips/0082LLB0W23}, and TCP~\cite{yao2023tcp}.


\subsection{Comparison with Existing methods}
In this section, we compare the proposed SEP with the existing methods from four types of tasks.

\textbf{Base-to-New Class Generalization:} Similar to existing CoOp-based methods, a significant evaluation of prompt tuning is the base-to-new generalization setting consisting of two disjoint subsets: \emph{Base} and \emph{New} classes.
The base-to-new generalization can be treated as zero-shot learning, which infers the trainable parameter with the labeled \emph{Base} classes and is evaluated on the disjointed \emph{New} classes.
As shown in Table~\ref{tab:base2new}, the proposed SEP obtains the highest average performance in the term of Base/New/H, \emph{e.g.,} obtaining the Base/New/H of 85.98\%/76.49\%/80.96\%.
Since SEP can effectively inject the prior discriminative knowledge inheriting from the pre-trained visual tokens, it obtains the performance on \emph{Base} classes of 85.98\%, obtaining 1.72\% improvement upon existing SOTA PromptSRC.
The superior performance also demonstrates the necessity and the reasonable of considering the prior knowledge for improving the discriminative on the seen domain. 
Moreover, we can observe that the proposed SEP obtains the \emph{New} performance of 76. 49\%, demonstrating that the use of the self-enhanced prompt can improve the generalization in the unseen domain.
In conclusion, the superior performance of SEP shows that using the self-enhanced prompt with considering the self-pretrained tokens can boost discriminative and generalization embeddings.

\begin{table}
\center
\small
\caption{Comparison of domain generalization.}
\label{tab:dg}
\resizebox{\linewidth}{!}{
\begin{tabular}{l|ccccc|c}
\toprule
	& ImageNet & -V2 & -S & -A & -R & Avg.\\
\midrule
CoCoOp & 71.02 & 64.07 & 48.75 & 50.63& 76.18& 59.91 \\
ProGrad & 72.24 & 64.73 & 47.61 & 49.39 & 74.58 & 59.08 \\
KgCoOp & 71.2 & 64.1 & 48.97 & 50.69& 76.7 & 60.12 \\
MaPLe & 70.72 & 64.07 & 49.15 & 50.9 & 76.98 & 60.27\\
DAPT & 71.67 & 64.5 & 49.53 & 51.1 & 76.33 & 60.37 \\
SEP & 71.03 & 64.30 & 49.75 & 50.70 & 77.43 & 60.54 \\
\bottomrule
\end{tabular}}
\vspace{-1.5em}
\end{table}

\begin{table*}
\caption{Comparison of cross-dataset evaluation from the ImageNet to the rest Ten datasets.}
\vspace{-0.5em}
\label{tab:crossdataset}
\scriptsize
\centering
\resizebox{\linewidth}{!}{
\begin{tabular}{l|ccccccccccc|c}
\toprule
  Datasets            & CLIP  & CoOp  & ProGrad & KgCoOp & DePT           & VPT   & PLOT        & PromptSRC      & MaPLe   & DAPT           & TCP         &   SEP\\
\midrule
ImageNet      & 66.70  & 71.51 & 72.24   & 70.66  & 72.77          & 69.73 & 71.60        & 71.27          & 70.72   & 71.60  & 71.40          &  70.33\\
\midrule
Caltech101    & 93.30  & 93.70 & 91.52   & 93.92  & 94.23 & 93.67 & 92.07       & 93.60          & 93.53   & 93.50           & 93.97          &94.17 \\
Pets    & 89.10  & 89.14 & 89.64   & 89.83  & 90.03          & 89.27 & 90.10        & 90.25          & 90.49   & 90.67          & 91.25 & 90.77\\
Cars & 65.70  & 64.51 & 62.39   & 65.41  & 65.57          & 65.5  & 65.70        & 65.70          & 65.57   & 65.93 & 64.69          & 65.90\\
Flowers       & 70.70  & 68.71 & 67.87   & 70.01  & 70.57          & 70.2  & 69.23       & 70.25          & 72.20    & 71.70           & 71.21 & 71.93\\
Food101       & 85.90  & 85.30 & 85.40   & 86.36  & 86.37          & 86.27 & 86.23       & 86.15          & 86.20   & 86.10           & 86.69 & 86.15\\
Aircraft  & 24.90  & 18.47 & 20.16   & 22.51  & 23.27          & 22.13 & 25.00 & 23.90          & 24.74   & 23.03          & 23.45         & 24.77\\
SUN397        & 62.60  & 64.15 & 62.47   & 66.16  & 66.67          & 66.57 & 61.67       & 67.10          & 67.01   & 67.00             & 67.15 & 67.07\\
DTD           & 44.30  & 41.92 & 39.42   & 46.35  & 45.97          & 46.93 & 38.60        & 46.87 & 46.49   & 44.00             & 44.35          & 45.63\\
EuroSAT       & 48.30  & 46.39 & 43.46   & 46.04  & 43.53          & 47.43 & 47.83       & 45.50          & 48.06   & 52.47 & 51.45          & 50.97\\
UCF101        & 67.60  & 66.55 & 64.29   & 68.50  & 69.30 & 67.20  & 67.00          & 68.75          & 68.69   & 68.73          & 68.73        & 68.50\\
\midrule
Avg.          & 65.24 & 63.88 & 62.71   & 65.51  & 65.55          & 65.52 & 64.34      & 65.81          & 66.30   & 66.31 & 66.29         &\textbf{66.59}\\
\midrule 
\end{tabular}}
\vspace{-1.0em}
\end{table*}

\textbf{Domain Generalization:} Domain generalization is conducted between the imagenet and the variant of imagenets (ImageNetV2, ImageNet-Sketch, ImageNet-A, and ImageNet-R) to verify the generalization ability of the prompt tuning.
The comparison with existing methods is summarized in Table~\ref{tab:dg}.
It can be seen that the proposed SEP obtains the best average performance of 60.54\%, proving the generalizability of the proposed SEP.

\begin{table*}
\caption{Comparison of few-shot learning with 4-shot samples.}
\vspace{-0.5em}
\label{tab:fsl}
\small
\centering
\resizebox{\linewidth}{!}{
\begin{tabular}{l|cccccccccccc|c}
\toprule
              & CLIP  & CoOp  & CoCoOp & ProGrad & KgCoOp & MaPLe & TIP-Adapter & DAPT  & PromptSRC & PLOT  & TaskRes &  TCP  & SEP\\
\midrule
ImageNet      & 66.70  & 69.37 & 70.55  & 70.21   & 70.19  & 70.67 & 70.78         & 70.80  & \textbf{70.80}     & 70.40  & 62.87 &    70.48 & 69.93\\
Caltech101    & 93.30  & 94.44 & 94.98  & 94.93   & 94.65  & 94.30  & 94.77         & 94.23 & 94.77     & 95.13 & 94.67   &             95.00   & \textbf{95.40}\\
Pets    & 89.10  & 91.30 & 93.01  & 93.21   & 93.20  & 92.05 & 92.26         & 92.17 & \textbf{93.23}     & 92.55 & 92.00 &               91.90  &92.90\\
Cars & 65.70  & 72.73 & 69.10  & 71.75   & 71.98  & 68.70  & 74.42         & 74.40  & 71.83     & 76.25 & 75.90&                          76.30 &\textbf{77.73}\\
Flowers       & 70.70  & 91.14 & 82.56  & 89.98   & 90.69  & 80.80  & 92.98         & 92.37 & 91.31     & 92.93 & 91.50  &                   94.40  &\textbf{94.63}\\
Food101       & 85.90  & 82.58 & 86.64  & 85.77   & 86.59  & \textbf{86.90}  & 86.18         & 83.60  & 86.06     & 86.46 & 86.03 &           85.3  &85.77\\
Aircraft  & 24.90  & 33.18 & 30.87  & 32.93   & 32.47  & 29.03 & 35.49         & 32.47 & 32.80     & 35.29 & 33.80   &       36.20  &\textbf{38.07}\\
SUN397        & 62.60  & 70.13 & 70.50  & 71.17   & 71.79  & 71.47 & 70.65         & 72.20  &72.80     & 71.73 & 72.70  &     72.11 &\textbf{73.30}\\
DTD           & 44.30  & 58.57 & 54.79  & 57.72   & 58.31  & 54.73 & 61.70          & 61.37 & 60.64     & 62.43 & 59.57 &      63.97 &\textbf{66.23}\\
EuroSAT       & 48.30  & 68.62 & 63.83  & 70.84   & 71.06  & 54.87 & 78.27         & 72.73 & 75.02     & 83.21  & 72.87 &     77.43 &\textbf{85.20}\\
UCF101        & 67.60  & 77.41 & 74.99  & 77.82   & 78.40  & 73.70 & 79.73         & 79.40  & 79.35     & 79.76 & 76.10 &      80.83 &\textbf{81.90}\\
\midrule
Avg.          & 65.37 & 73.59 & 71.98  & 74.21   & 74.48  & 70.66 & 76.11         & 75.07 & 75.33     & 76.9& 74.36 &   76.72 & \textbf{78.28}\\
\bottomrule
\end{tabular}}
\vspace{-1.5em}
\end{table*}

\textbf{Cross-Dataset Generalization:} Different from the base-to-new generalization assuming that the testing dataset and training dataset have a similar data distribution with disjoint classes, the cross-dataset generalization performs the training on the base dataset (ImageNet) and evaluation on the rest ten datasets.
The comparison between SEP and the existing methods is summarized in Table~\ref{tab:crossdataset}.
We can observe that the proposed SEP obtains the highest average performance, \emph{e.g.,} obtaining the average performance of 66.59\%.
The superior performance demonstrates the effectiveness of SEP in generating the generalization description by injecting the self-pretrained token knowledge.


\textbf{Few-Shot Learning:} To further verify the effectiveness of the proposed SEP, we conduct the few-shot learning with K-shot labeled images.
As shown in Table~\ref{tab:fsl}, the proposed SEP obtains an obvious improvement over existing methods.
For example, SEP obtains the average performance of 78.28\%, obtaining an improvement of 1.38\% over the existing PLOT.
Moreover, SEP obtains the highest performance in most datasets, \emph{e.g.,} 8/11 datasets.
The reason is that the self-enhanced prompt after injecting the prior discriminative knowledge of the pre-trained tokens can provide significant clues. 
Especially for the few-show learning that can provide less useful knowledge, more critical knowledge can be provided by the self-enhanced prompt for inferring discriminative knowledge.

\subsection{Ablation Study}
We conduct several ablation studies to verify the effectiveness of SEP.

\begin{table}
\center
\caption{\small Effect of Self-enhanced Prompt tuning.}
\label{tab:sep}
\small
\begin{tabular}{cc|ccc}
\toprule
Visual Encoder & Text Encoder & Base & New & H \\
\midrule
IVLP&IVLP&85.91&74.99&80.08\\
IVLP&	SEP	&86.19&75.35&80.41\\
SEP&	IVLP	&85.53&74.80& 79.80\\
SEP&	SEP	&85.98&76.49&80.96\\
\bottomrule
\end{tabular}
\end{table}

\begin{table*}[htb]
\small
\caption{\small Effect of inserting SEP into different layers.}
\label{tab:layers}
\resizebox{0.9\linewidth}{!}{
\begin{tabular}{c|ccc|cccc|c}
\toprule
Layers & \{4\} & \{8\}& \{10\} & \{4,8\} & \{4,8,10\} & \{3,6,9\} & \{2,4,6,8,10\}& \{1,2,3,4,5,6,7,8,9,10,11\}\\
\midrule
Base  & 84.66	&84.62	&85.16	&85.38	&85.44	&85.32	&\textbf{86.27}	&85.98	\\
New   & 74.8	&76.07	&75.34	&75.19	&75.41	&74.95	&75.92	&\textbf{76.49}	\\
H     & 79.43	&80.12	&79.95	&79.96	&80.11	&79.80	&80.76	&\textbf{80.96}\\
\bottomrule
\end{tabular}}
\vspace{-1.0em}
\end{table*}

\noindent\textbf{Effect of Self-enhanced Prompt tuning:}
To verify the effectiveness of Self-enhanced Prompt tuning(SEP), we replace SEP with the Individual Visual-Language Prompting (IVLP)~\cite{MAPLE23} that conducts the individual input-independent prompt tuning.
As shown in Table~\ref{tab:sep}, based on the baseline with conducting IVLP on visual prompt tuning, using the SEP for textual prompt tuning obtains a higher performance on all three types of measure metrics, \emph{e.g.,} improves the Base/New/H from 85.91\%/74.99\%/80.08\% to 86.19\%/75.35\%/80.41\%. 
Moreover, conducting SEP on both visual and textual prompt tuning achieves the performance of 76.49\%/80.96\% for New/H, obtaining 1.5\%/0.9\% improvement upon the ones using IVLP on both modals. 
The superior performance demonstrates the effectiveness of the proposed self-enhanced prompt tuning.

\noindent\textbf{Insert Which Layer:}
As the encoders in CLIP consist of multiple layers, the effect of inserting SEP into different layers is analyzed and summarized in Table~\ref{tab:layers}.
The best results were achieved by inserting the prompts into all layers.

\noindent\textbf{Effect of Multi-model prompt tuning:}
SEP can be easily applied for visual prompt tuning and textual prompt tuning for Visual and Text Encoders.
We thus analyze the effect of the multi-modal prompt tuning and summarize the results in Table~\ref{tab:MMPT}.
It can be observed that conducting SEP on two modals obtains a higher performance than merely one modal, proving that the two modal prompt tuning are complementary and conducting the multi-modal prompt tuning is more effective.

\begin{table}
\caption{ Effect of Multi-modal Prompt tuning.}
\label{tab:MMPT}
\begin{tabular}{cc|ccc}
\toprule
Visual Encoder & Text Encoder & Base & New & H \\
\midrule
-&-&82.86&74.30&78.35\\
-&	SEP	& 84.66&75.51&79.82\\
SEP& -&82.94&75.17&78.86\\
SEP&	SEP	&85.98&76.49&80.96\\
\bottomrule
\end{tabular}
\end{table}

\begin{table}
    \centering
    \small
    \caption{Effect of Token Selective Strategy.}
    \label{tab:rts}
    \begin{tabular}{cc|ccc}
    \toprule
    Visual Encoder & Text Encoder & Base & New & H \\
    \midrule
    Activation&	Front	& \textbf{85.98}&76.49&\textbf{80.96}\\
    Front&	Front	& 85.8&\textbf{76.5}&80.88\\
    Front& Activation&85.31&75.84&80.30\\
    Activation&	Activation	&85.38&75.70&80.25\\
    \bottomrule
    \end{tabular}
\end{table}

\noindent\textbf{Effect of Token Selective Strategy:} There are two types of strategy to choose representative tokens from all pre-trained tokens: front-based and activation-based selective strategies, where the front-based strategy selects the front of $L_v$ tokens, and the activation-based strategy selects the tokens with top-$L_v$ high activation scores. 
As shown in Table~\ref{tab:rts}, the front-based strategy is suitable for textual prompt tuning, and the activation-based strategy is suitable for visual prompt tuning.

\noindent\textbf{Effect of Token Fusion Strategy:} 
After selecting the representive tokens from all tokens, Token Fusion Module(TFM) applies the cross-attention to fuse the selected tokens and the prompt-related tokens.
We further make comparion with other two token fusion strategy: Add and MLP strategies, where `Add' is the add operation, and `MLP' denotes use multiple fully-connected layers to fuse.
As shown in Table~\ref{tab:tfs}, TFM with cross-attention obtains the best performance.

\begin{table}
    \centering
    \small
    \caption{Effect of Token Fusion Strategy.}
    \label{tab:tfs}
    \begin{tabular}{c|ccc}
    \toprule
    Fusion Strategies & Base & New & H \\
    \midrule
    	Add	&83.39&74.09&78.46\\
  	MLP	&85.78&75.77& 80.46\\
     TFM 	&\textbf{85.98}&\textbf{76.49}&\textbf{80.96}\\
    \bottomrule
    \end{tabular}
\vspace{-1.0em}
\end{table}

\begin{figure}
\includegraphics[width=0.9\linewidth]{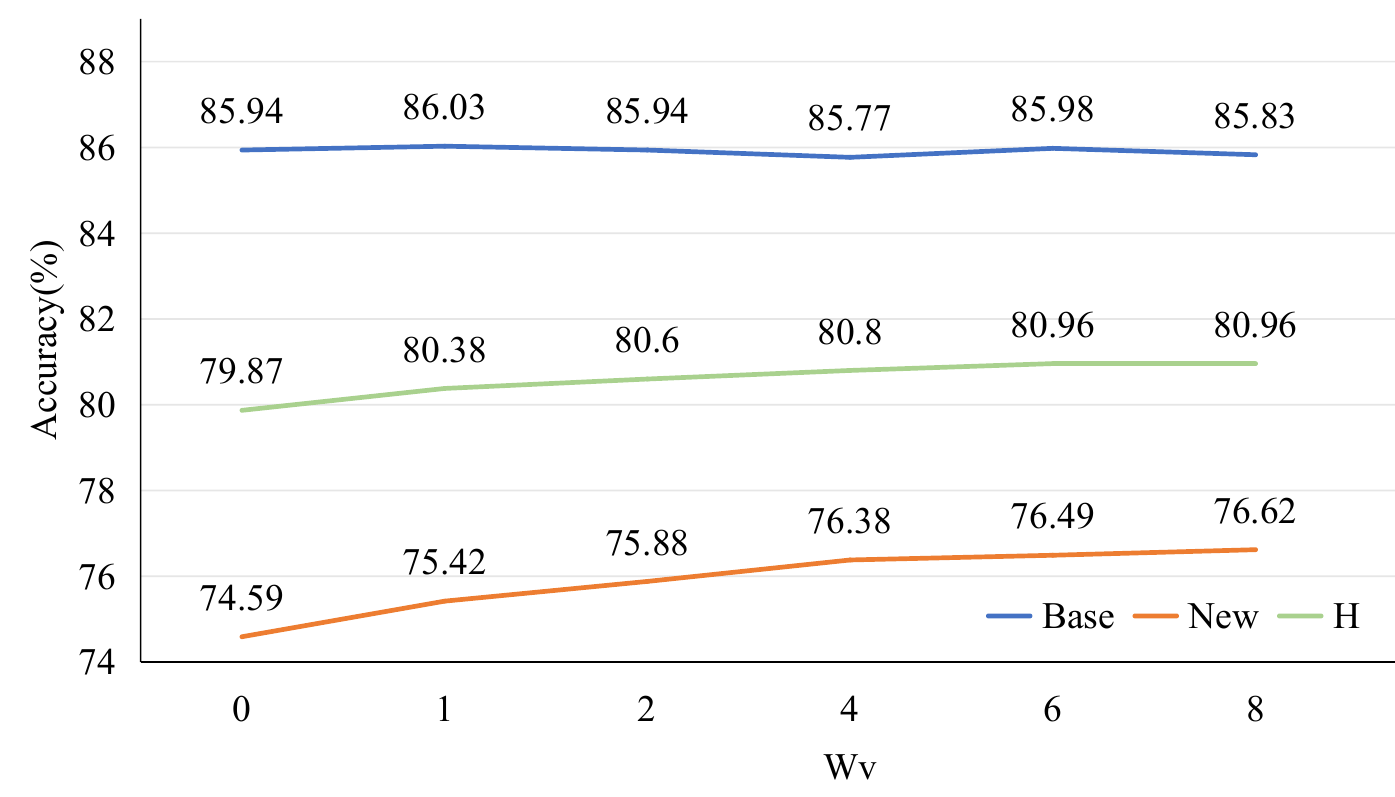}
\caption{Effect of $W_v$.}
\label{fig:wv}
\vspace{-1.0em}
\end{figure}

\noindent\textbf{Effect of $\omega_v$:}
The effect of the hyperparameter $\omega_v$ in the final objective(Eq.~\eqref{eq:final}) is shown in Figure~\ref{fig:wv}.
It can be observed that firstly increasing the hyperparameter would increase the performance while degrading the performance.
Consequently, setting $\omega_v$=6.0 obtains the best performance.

\section{Conclusion}
To overcome the shortcoming that existing CoOp-based methods have the limited ability to capture the input-aware specific knowledge, we propose a novel Self-Enhanced Prompt tuning by utilizing the Token Fusion Module(TFM) to inject the discriminative knowledge contained in self-pretrained tokens into the self-enhanced prompts.
Evaluation of several tasks and benchmarks verify the effectiveness of the proposed SEP for prompt tuning.
In the proposed SEP, the choice of representative tokens is still an open and critical problem to improve the robustness and flexibility of the SEP.
{
    \small
    \bibliographystyle{ieeenat_fullname}
    \bibliography{egbib}
}


\end{document}